\documentclass{article}
\usepackage{ijcai/ijcai21}

\usepackage{times}
\usepackage{soul}
\usepackage{url}
\usepackage[hidelinks]{hyperref}
\usepackage[utf8]{inputenc}
\usepackage[small]{caption}
\usepackage{graphicx}
\usepackage{amsmath}
\usepackage{amsthm}
\usepackage{booktabs}

\usepackage{algorithm}
\usepackage{tabularx}
\usepackage{booktabs}
\usepackage{comment}
\usepackage[noend]{algpseudocode}
\usepackage{xcolor}
\usepackage[switch]{lineno}
\usepackage{multicol}
\usepackage{multirow}
\usepackage{graphicx}
\usepackage{subcaption}
\usepackage{amsfonts}

\makeatletter
\newcommand{\multiline}[1]{%
  \begin{tabularx}{\dimexpr\linewidth-\ALG@thistlm}[t]{@{}X@{}}
    #1
  \end{tabularx}
}
\makeatother

\title{EARN: A Multi-Objective Evolutionary Method \\for Generating Efficient DNN-Ensembles}

\author{
Marc Ortiz$^1$\footnote{Contact Author}\and
Florian Scheidegger$^2$\and
Marc Casas$^1$\and
Cristiano Malossi$^2$\And
Eduard Ayguade$^1$
\affiliations
$^1$Barcelona Supercomputing Center\\
$^2$IBM Research - Zurich, Switzerland\\
}

\begin{document}

\maketitle
 
\begin{abstract}

In this work, we leverage ensemble learning as a tool for the creation of faster, smaller, and more accurate deep learning models. We demonstrate that we can jointly optimize for accuracy, inference time, and the number of parameters by combining DNN classifiers. To achieve this, we consider multiple ensemble strategies: bagging, boosting, and an ordered chain of classifiers. To reduce the number of evaluations during the search of efficient ensembles, we propose EARN, an evolutionary approach that optimizes the ensemble regarding specific constraints. We run EARN on 10 image classification datasets on both CPU and GPU platforms, and we generate models with speedups up to $4.07\times$ on a GPU device, reductions of parameters by $7.70\times$, or increases in accuracy up to $7.30\%$ regarding the best DNN in the pool. Additionally, our method generates models that are $4.70\times$ smaller than state-of-the-art methods for automatic model generation.

\end{abstract}
\section{Introduction}

Over the past years, mainly due to significant improvements in terms of training data quality, computational capacity, and efficient learning techniques, Deep Learning (DL) methods have achieved outstanding results in a wide range of applications including object recognition, scene understanding, speech recognition, language processing, or motion planning. Despite this success, a costly and long training process is generally required to generate accurate DL models. In addition, these models must meet strong restrictions in terms of memory footprint, latency, prediction performance, or energy. 

Ensemble Learning (EL) is a very popular area in Machine Learning (ML) that combines multiple learners on a single-learning task. EL is aimed at generating composed models with better properties than the individual building blocks. It has been experimentally and theoretically demonstrated that if the individual learners are accurate and diverse, some ensemble models with higher predictive performance than any single learner can be obtained~\cite{58871}. The potential of ensemble learning methods to generate efficient solutions have been demonstrated. For example, a random forest~\cite{breiman2001random} combines multiple decision trees with bagging to obtain a combined solution with high classification accuracy and low inference response time. Previous work~\cite{viola2004robust} accelerates a face detector by arranging the classifiers in the ensemble in a cascade way with increasing order of complexity, and performing an early-exit condition when a clear non-face patch of the image is encountered.

The following steps are key for the creation of efficient ensembled solutions: The \textit{generation} of a pool of learners, the \textit{selection} of subset learners, and finally the \textit{combination/interaction} of the learners. If the pool of learners is composed of homogeneous machine learning models, some techniques are applied to introduce diversity among the learners \cite{breiman1996bagging,freund2001adaptive}. Otherwise, a diverse pool of Pareto optimal heterogeneous learners is generated with some multi-objective optimization algorithm~\cite{sopov2015self,chandra2004divace}. Different optimization algorithms have been explored for the subset selection, e.g., greedy \cite{nan2012diversity}, evolutionary \cite{qian2015subset} or semi-definite programming \cite{zhang2006ensemble}.  Although evolutionary methods can be costly, they allow to find global optimal solutions, and are widely adopted for solving multi-objective optimization problems. Finally, a voting protocol usually fusions the predictions of all the selected learners. 

State-of-the-art techniques present some limitations that our method addresses. First, while previous approaches just consider a single ensemble strategy, we demonstrate in this work that much better ensembles can be constructed by considering multiple ensemble strategies. Thus, an ensemble is not a bag of learners anymore, but rather a graph where components interact with each other to provide a single merged solution. Our evolutionary method is not just mutation-driven like many previous methods, since it can apply crossover operations on the graph to improve the search. In addition, most methods adopt an abstract representation of the cost of running the ensemble, such as the number of models in the ensemble, or the number of instances executed per model. This does not work when learners are heterogeneous and does not take into account the hardware platform where models run. Our approach is fully aware of the model's latency on the targeted architecture platform, which provides very large improvements with respect to the state-of-the-art. 
\\

\begin{figure*}[!tbp]
  \centering
  \begin{minipage}[b]{0.32\textwidth}
    \includegraphics[width=\textwidth]{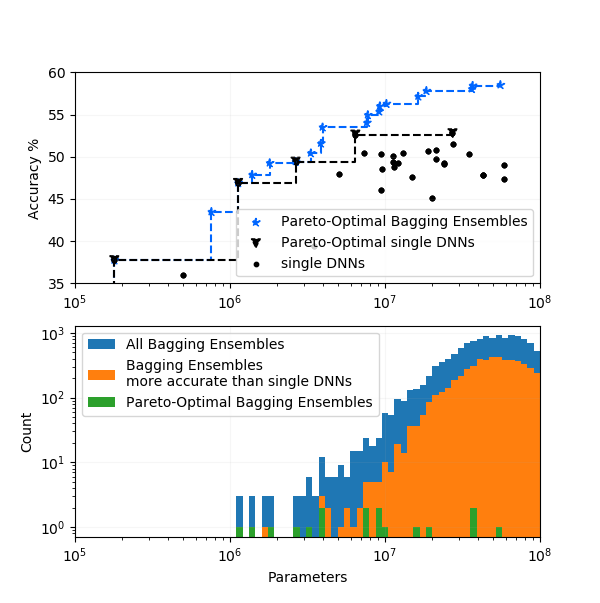}
    \subcaption{Bagging of 3 DNNs}
    \label{fig:motivation_bagging}
  \end{minipage}
  \begin{minipage}[b]{0.32\textwidth}
    \includegraphics[width=\textwidth]{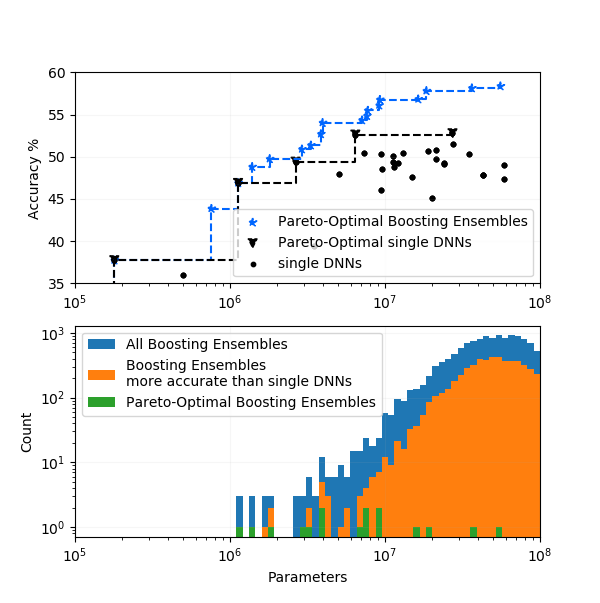}
    \subcaption{Boosting of 3 DNNs}
    \label{fig:motivation_boosting}
  \end{minipage}
  \begin{minipage}[b]{0.32\textwidth}
    \includegraphics[width=\textwidth]{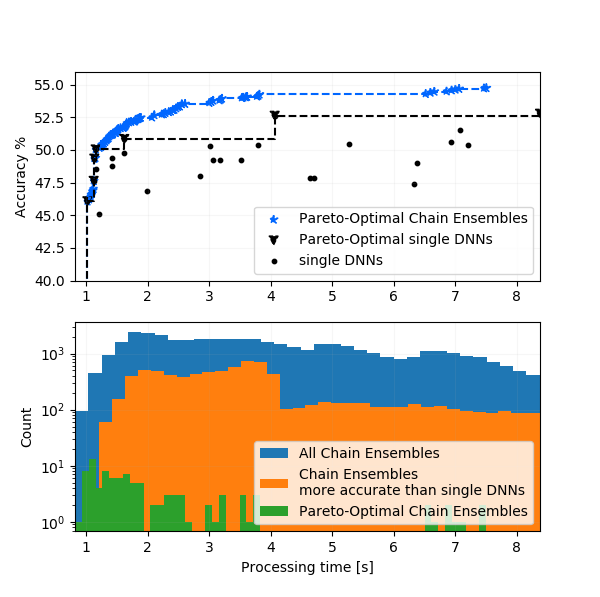}
    \subcaption{Chaining of 2 DNNs}
    \label{fig:motivation_chain}
  \end{minipage}
  \caption{Improving the performance vs cost trade-offs drawn by an initial pool of  32 DNNs trained on \textbf{Caltech-256}.}
\end{figure*}

In particular, we make the following contributions beyond the state-of-the-art:
\begin{itemize}
    \item We combine bagging, boosting, and ordered chain of classifiers to generate efficient ensembles of Deep Neural Networks (DNNs). Leveraging distinct ensemble methodologies enables to simultaneously optimize classification accuracy, inference time, and model size.
    \item We propose EARN, a multi-objective evolutionary algorithm for the automatic generation of optimal ensembles. EARN requires very few evaluations to generate close to optimal solutions. Unlike similar ensemble optimization proposals, EARN leverages multiple ensemble learning techniques, constructs heterogeneous DNN topologies, considers multiple model combinations besides mutation, and optimizes the ensemble for the targeted hardware platform. We evaluate EARN on both CPU and GPU architectures to demonstrate it automatically produces models that are 4.70$\times$ smaller than the state-of-the-art, and 4.07$\times$ times faster than the best individual DNN building block. EARN also achieves large improvements in terms of accuracy and model size, with up to 7.30\% and $7.70\times$ respectively. 
\end{itemize}
\section{Background and Motivation}
This section describes some state-of-the-art approaches to combine classifiers, and justifies the need for an automatic approach to obtain the best combinations.
\subsection{Bagging, Boosting and Chain of Classifiers}
Bootstrap aggregating also known as \textbf{bagging} is composed of two steps. First, a fixed-sized and stochastically-sampled subset $D_i$ of a training dataset $D = \{ (x,l) : x \in \mathbb{R}^d,  l \in \mathbb{R} \}$ is fed to each of the learners to encourage diversity in their predictions. Second, during the aggregation step, the predictions of the individual models $h_i(x): x \in \mathbb{R}^d \xrightarrow{} \mathbb{R}$ in the ensemble $H(x)$ are merged to obtain a single solution. We consider three merging protocols: \textit{average}, \textit{voting}, and \textit{max}. This paper considers a heterogeneous set of learners, as Section~\ref{sec:experimentalsetup} shows. Since we already have a large degree of diversity, we avoid performing the aggregation step.

\textbf{Boosting} considers a succession of models $h_i(x)$ and sequentially trains them to mitigate the error of the previously used $h_{0..i-1}$ models. Boosting simultaneously reduces both variance and bias errors. A weight $\sigma$ is assigned to each classifier regarding its performance on the train set. Those weights are later used to merge the predictions. In this work we consider \textit{weighted average}, \textit{weighted voting}, and \textit{weighted max} as merging methods.
We consider SAMME \cite{hastie2009multi}, which is a multi-class variation of the well-known AdaBoost algorithm~\cite{freund1996experiments}. Moreover, due to the training costs and the large number of DNN boosted combinations that can arise, we modify SAMME so that, like in bagging, we train DNNs one time in parallel.

In an \textbf{ordered chain of classifiers}, there is an order of models execution in the ensemble $H(x)$, and an early-exit condition to speed up the inference response. It is a form of dynamic ensemble pruning.  Some approaches~\cite{wang2018quit} leverage an ordered chain of binary classifiers with early-stopping thresholds to reduce inference time and maintain the classification accuracy. On average, they obtain speedups of 2$\cdot$-4$\cdot$ regarding the inference with the whole chain. Additionally, since threshold values are task-dependent, previous work~\cite{inoue2019adaptive} defines a statistically rigorous exit condition based on confidence intervals.

Inspired by the previous approaches, we consider a chain of classifiers where the early-exit condition is controlled by the threshold $\tau\in[0..1]$. During the inference process, we compare the highest activation value of the DNN's $Softmax(z)$ to $\tau$. If the activation value is lower than $\tau$, we query the next classifier in the chain. The inference process can be described with the following equation, starting at $i=0$:

\begin{equation}
    h_i(x) = 
    \begin{cases}
        h_{i+1}(x) & \text{if } \tau_i \geq max(h_i(x))\\
        h_i(x) & \text{otherwise}
    \end{cases}
\end{equation}

Techniques based on either Bagging, Boosting or Chains of Classifiers display complex trade-offs in terms of accuracy versus model size, among others. In addition, they display a large variety of behaviors depending on the merging policy. The next section presents some examples of such complex search space, which justify the need for an automatic and fast method to navigate across the huge space of combinations.

\subsection{Analysis of Ensemble Methods}
We evaluate all possible bagging and boosting ensembles of 3 models from a pool of 32 DNN models trained on the Caltech-256 dataset. 
The pool is a diverse set of models with different complexities including many state-of-the-art architectures for image-related tasks.
Section~\ref{sec:experimentalsetup} describes in detail our experimental setup.
For each bagging and boosted ensemble, we consider 3 merging protocols: \textit{(weighted) average, (weighted) voting, and (weighted) max}. 

Figure~\ref{fig:motivation_bagging} displays the best bagging combinations plus all the 32 individual classifiers in their top plots. The x-axis shows the size of the solutions in terms of the number of parameters, while the y-axis displays test accuracy. Accuracy and model size show a correlation since large models display better accuracy. There is no single optimal solution for the Caltech-256 problem, but a set of optimal solutions that constitute a Pareto-optimal curve. A black dotted line unifies the optimal single-DNN models, and a blue dotted line unifies the optimal boosting ensembles. Many of the boosted ensembles display high accuracy properties, and several outperform optimal single DNNs. In addition, the number of Pareto-optimal ensemble solutions is larger than single-DNN models, which allows to target a wide range of scenarios in terms of accuracy or memory footprint. 

Figure~\ref{fig:motivation_bagging} also displays the total number of possible bagging ensembles for a certain parameter count generated from the initial set of 32 DNNs. In blue, we represent all possible ensembles, in orange the ensembles that improve the accuracy of the most accurate single DNNs within that parameter limit, and in green the Pareto-Optimal ensembles. Only a tiny fraction of the ensembles belong to the Pareto-Optimal set. For some parameter counts there is very large number of possible combinations, which illustrates the impossibility of performing an exhaustive search. Figure~\ref{fig:motivation_boosting} shows our evaluation considering boosting, which displays very similar results as bagging.

Analogous to the study of bagging and boosting, we evaluate all possible chains of length 2 by combining pairs of DNNs from the pool of 32 models. For every pair, we generate 100 chains by modifying the threshold from 0 to 1 with steps of 0.01 and placing the smallest DNN in the first chain position. We display the optimal chains and the individual DNNs in the top of Figure~\ref{fig:motivation_chain}. The y-axis of the figure shows the test accuracy, while the x-axis represents the seconds required for the solutions to process $10^4$ 32x32 test images on Caltech-256, with the GPU described in Section~\ref{sec:experimentalsetup} and batches of 128 samples. Some chains outperform the single-DNN predictors in terms of inference time and test accuracy. Fast and accurate solutions can be designed because the complex DNN in the chain is only triggered for hard to classify samples, while the smallest model is activated by default and solves easy cases. Lower threshold values produce faster solutions while higher threshold values produce slower but more accurate solutions. Figure~\ref{fig:motivation_chain} shows that a small number of combinations improve single-DNN models. 

Even when considering simple combinations of a few DNNs, exhaustively searching for the best ensembles requires exploring a very large number of configurations. 
In addition, just a small subset of all possible ensemble combinations provides solutions that significantly improve the original set of single-DNN models in terms of execution time, accuracy, or size.
If we consider complex ensembles of DNNs combining bagging, boosting, and chaining to jointly optimize for accuracy, inference time, and model size the search space will be even larger than just considering chain, boosting, or bagging ensembles alone. 


The automatic methodology we propose in the following section makes it possible to generate this set of optimal ensemble solutions without incurring the huge overhead of randomly generating combinations of models. In addition, the flexibility that we enable by considering a wide range of model combinations, makes it possible to automatically generate optimal ensembles according to multiple objectives.

\section{EARN}
In this section, we present an Evolutionary Approach for Reducing the Number of ensembles to evaluate (EARN) that automatically generates efficient DNN ensembles. We chose evolutionary algorithms as the optimization method due to their simplicity, computational scalability, and record of success in the Multi-Objective Optimization (MOO) field \cite{deb2002fast,zitzler2001spea2}. EARN follows the Genetic Programming (GP) framework \cite{koza1992genetic} in the representation of individuals and genetic operators. EARN also follows the NGSA-II algorithm \cite{deb2002fast}, performing an efficient discovery and mainteinance of a diverse set of Pareto-optimal ensembles.

\begin{figure}
\begin{center}
\includegraphics[width=8cm]{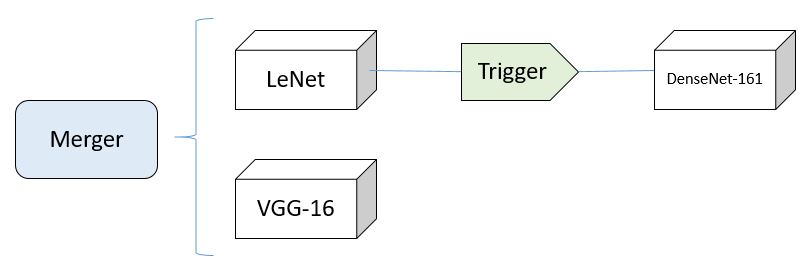}
\end{center}
\caption{Bagging/Boosting a chain of DNNs, and a single DNN.}
\label{fig:individuals_earn}
\end{figure}

\textbf{Population.} {The population is composed of a set of $M$ ensembles. We represent the ensembles as Direct Acyclic Graphs (DAG). Nodes in the DAG can be split into three categories: classifier, merger, or trigger nodes. A classifier node receives a set of inputs and produces predictions for each input. Even though we employ DNNs as our classification algorithm, classifier nodes can be instantiated with any other ML model.
Trigger nodes decide which classifiers to activate for each input. We employ an analytical trigger with a threshold value to discriminate between well-classified and weakly-classified inputs based on the classifier output. The trigger forwards the weakly-classified inputs to the next classifiers in the chain. Our trigger design is simple although it supports more complex triggering policies.
Finally, the merger node combines multiple predictions to produce a single unified prediction. 
Figure \ref{fig:individuals_earn} shows an example of an ensemble that merges the outputs of a 2-DNN chain with a 1-DNN predictor. It has three classifier nodes, one trigger node and one merger node}.

\textbf{Population sets.} The initial population are ensembles composed of a single classifier. We represent this initial set as $P$. We denote the offspring produced by EARN as $O$.

\textbf{Mutation.} As in GP, EARN performs mutation operations directly to the population of ensembles $P$ with a frequency of $0 \leq m_r \leq 1$. A mutation on an individual $s$ creates a new individual $s'$ that is later added to the offspring $O$.
When an ensemble mutates, one or more nodes in the graph of the ensemble change. The parameter $m_p$ defines the probability of nodes to mutate. If a \textit{merger} mutates, we either 1) add another DNN selected at random to the bagging/boosting ensemble, or 2) switch the current merging protocol between average (bagging) or weighted average (boosting). If a \textit{trigger} mutates, we 3) increment or decrement (decided at random) the threshold by a small constant of $0.1$. If a \textit{classifier} mutates, we 4) replace it with another DNN from the pool of DNNs selected at random, or 5) extend the chain by attaching a pair of trigger-classifier. A new trigger is initialized with a threshold value of $\tau=0.5$.

\textbf{Crossover.} EARN selects two parent ensembles $s_1$ and $s_2$ from $P$, and it performs a single-point crossover operation on them to create two more ensembles $s_1'$,$s_2'$, which are added to the offspring $O$. The frequency of the crossover operation is controlled by the rate of crossover parameter $c_r=1-m_r$. The $C$ parameter limits the offspring spawned from mutation and crossover operations at each generation.

\textbf{Fitness.}
Like in the NGSA-II algorithm, two values define the fitness of an ensemble solution when optimizing $n$ objective functions $y_i$, the \textit{non-dominated rank}, and the \textit{crowding distance}. The \textit{non-dominated rank} of a solution $s$ is the number of solutions that dominate $s$. A solution $s_1$ dominates another $s_2$ if $ \forall i\in\{1..n\};$ $y_i(s_1)$ is not worst than $y_i(s_2)$, and $\exists i$ such that $y_i(s_1)$ is better than $y_i(s_2)$. Solutions with the same rank, belong to the same front. The solutions whose rank is 0 are the Pareto-optimal solutions, and belong to the Pareto-optimal front. Then, the \textit{crowding distance} of a solution $s$ measures how unique is this solution, or how far is this solution from other solutions within the same front. Before computing the crowding distance, objective values should be normalized. A solution $s_1$ is fitter than $s_2$ if the \textit{non-dominated rank} of $s_1 < $ rank $s_2$. If rank $s_1 = $  rank $s_2$, the solution with highest \textit{crowding distance} wins. This way, we simultaneously improve the Pareto-optimal frontier of EARN's population, while maintaining the diversity of solutions across all solution space.

\begin{algorithm} 
\caption{: Sequential Execution of EARN}
\small
\label{algorithm:earn}
    \begin{algorithmic}[0]
     \Function{EARN}{}
        \State \multiline{%
            \textbf{$\downarrow$ In}: some trained $DNNs$; $M$ the population limit; $C$ the offspring limit;  $c_r$ crossover rate; $m_p$ probability of mutation; $K$ tournament size;}
        \State \multiline{%
            \textbf{$\uparrow$ Out}: $P$ the last generation of ensembles, $F$ the corresponding fitting values}
        \State $P \leftarrow DNNs$
        \State $F \gets$ \Call{fast\_nondominated\_sort}{$P$}
        \While{ \textbf{not} termination}
            \State $O \gets$ \Call{spawn\_offspring}{$P$, $F$, $K$, $C$, $c_r$, $m_p$}
            \State $F_o \gets$ \Call{fast\_nondominated\_sort}{$O$}
            \State $P$, $F \gets$ \Call{selection}{$P+O$, $F+F_o$, $M$}
        \EndWhile
    \EndFunction\\
    
    \Function{Spawn\_Offspring}{}
        \State \textbf{$\downarrow$ In}: $P$, $F$, $K$, $C$, $c_r$, $m_p$
        \State \textbf{$\uparrow$ Out}: $O$
        \State $O \leftarrow \{\}$
        \While{$\|O\| < C$}
            \State With probability $m_r$
            \State $O \leftarrow  O \cup \{ \Call{mutation}{P,F,K,m_p} \}$ 
            \State With probability $1-m_r$
            \State $O \leftarrow  O \cup \{\Call{crossover}{P,F,K}\}$
        \EndWhile
    \EndFunction\\
    \end{algorithmic}
\end{algorithm}

\textbf{Selection.} A \textit{tournament} selection mechanism, with tournament size of $K$, is employed for parent selection in crossover and individual selection in mutation.
A \textit{most-fit} selection method selects the $M$ best individuals in $P$ and $O$ according to the fitness. The $M$ selected individuals are alive next generation.

\textbf{Termination.} EARN allows to specify a termination criteria in terms of either a maximum number of iterations $I$, or  with the \textit{hypervolume indicator} \cite{zitzler1999multiobjective}. In this last case, the algorithm would stop after several iterations without observing a significant improvement in the Pareto-optimal front of the population of ensembles.

\textbf{EARN algorithm.} Algorithm~\ref{algorithm:earn} shows a high-level representation of EARN. The algorithm uses an initial set of single DNNs as its initial population $P$, and their corresponding fit values $F$. We follow NGSA-II's fast non-dominated sorting algorithm to compute the fitness values.
In each iteration, EARN produces offspring $O$ using the operations of crossover and mutation rates with values $c_r$ and $m_r$, respectively.
EARN selects the $M$ most fit individuals from the population $P$ and offspring $O$, and proceeds to the next iteration. At termination, EARN returns the alive population of ensembles, which includes the Pareto-optimal front.

\section{Evaluation}

    We evaluate the performance of EARN under two scenarios. Firstly, we showcase the consistency of our approach towards optimizing the ensembles according to multiple datasets and hardware platforms. In the second scenario, we demonstrate that EARN improves DL models automatically generated by state-of-the-art approaches. In both scenarios, we set EARN parameters as follows: Population limit $M$ to $500$, offspring limit $C$ to $200$, tournament size $K$ to $10$, the mutation rate $m_r$ to $0.4$, the probability of mutation $m_p$ to $0.6$, and the number of iterations $I$ to 100. This parameter setup causes the total number of evaluated ensembles to be 20,000. On the CPU system described in Section~\ref{sec:experimentalsetup}, EARN takes less than 25 hours to go through the $100$ iterations. Different parameter setups produce similar ensembles as long as they do not contain any unreasonably large or small parameters. We run EARN on validation sets. We use a different a data set, the test set, to evaluate its accuracy.


\textbf{EARN on 10 image data sets. } 
 We evaluate EARN on 10 datasets and the 2 computing systems described in Section~\ref{sec:experimentalsetup}. For each dataset and computing system, EARN is fed with an initial pool of 32 DNNs to generate enhanced combined solutions. In this analysis, we set EARN's optimization objectives to classification error, prediction time, and model size. Since the prediction response time of the same DNN solution changes between computing systems, the optimal ensembles found by EARN for one platform may differ from the best ensembles of the other platform. The Experimental Setup section provides details on such datasets, computing systems, and the pool of 32 DNNs.

Table \ref{table:earn_bigtable} displays the improvements achieved by EARN. In each dataset, we reveal the most accurate DNN in the pool (reference DNN), along with 3 optimal ensembles found by EARN. The \textit{ensemble~A} holds the highest accuracy, the \textit{ensemble~C} is the smallest ensemble surpassing the reference DNN's accuracy, and \textit{ensemble~B} is the fastest ensemble while also enhancing the reference's accuracy. As said previously, since the computing time of the DNN models varies with different hardware platforms, we provide two \textit{ensemble~B} solutions, the optimal for CPU and the optimal for a GPU architecture. Since EARN returns a set of Pareto-optimal ensembles at the end of its execution, \textit{ensembles~A}, \textit{B}, and \textit{C} are found on a single EARN run. For every optimal ensemble, we display three properties in the table: increase in accuracy (Acc.), speedup (Sp.), and size reduction (Size). The speedup is always measured on GPU apart from the results of the \textit{ensemble~B} (CPU).

We improve the accuracy of the reference DNNs on almost all datasets, with a top performance of 7.30\% increase. We also find faster solutions, providing speedups up to 4.07\% on GPU regarding the reference DNN. Finally, we reduce the size of the reference DNNs in all but one dataset. For some scenarios we deliver solutions close to one order of magnitude smaller than most accurate single-DNN model.

\begin{table*}[] 
\centering
\caption{EARN results on 10 image classification datasets, realtive to the reference DNN.}
\label{table:earn_bigtable}
\small
    \begin{tabular}{@{}ll@{\hspace{0.2cm}}ll@{\hspace{0.2cm}}l@{\hspace{0.1cm}}ll@{\hspace{0.2cm}}l@{\hspace{0.1cm}}ll@{\hspace{0.2cm}}l@{\hspace{0.1cm}}ll@{\hspace{0.2cm}}l@{\hspace{0.1cm}}ll@{}} 
    \toprule
    
    \multirow{2}{*}[-1mm]{\textbf{Dataset}} & 
    \multicolumn{2}{c}{\textbf{Reference DNN}} & 
    \multicolumn{3}{c}{\textbf{Ensemble A}} & 
    \multicolumn{3}{c}{\textbf{Ensemble B}} & 
    \multicolumn{3}{c}{\textbf{Ensemble B (CPU)}} & 
    \multicolumn{3}{c}{\textbf{Ensemble C}}\\ 
    \cmidrule(lr){2-3} \cmidrule(lr){4-6} \cmidrule(lr){7-9} \cmidrule(lr){10-12}\cmidrule(lr){13-15} 
    
    &  Acc. & Name &  Acc. & Sp. & Size &  Acc. & Sp. & Size &  Acc. & Sp. & Size & Acc. & Sp. & Size\\ 
    \midrule
     \textit{cifar10} &
    94.74{\scriptsize \% }& {\scriptsize DenseNet-161} & +0,56{\scriptsize \% }& 0,62{\scriptsize {\tiny $\times$ } } & 2,95{\scriptsize {\tiny $\times$ } } & +0,09{\scriptsize \% }& 3,66{\scriptsize {\tiny $\times$ } } & 1,47{\scriptsize {\tiny $\times$ } } & +0,16{\scriptsize \% } & 5,91{\scriptsize {\tiny $\times$ } } & 3,60{\scriptsize {\tiny $\times$}} & +0,06{\scriptsize \% }& 1,65{\scriptsize {\tiny $\times$ } } & 0,35{\scriptsize {\tiny $\times$ } } \\
    \textit{cifar100} &
    77.55{\scriptsize \% }& {\scriptsize DenseNet-161} & +2,59{\scriptsize \% }& 0,38{\scriptsize {\tiny $\times$ } } & 3,13{\scriptsize {\tiny $\times$ } } & +0,18{\scriptsize \% }& 2,36{\scriptsize {\tiny $\times$ } } & 1,35{\scriptsize {\tiny $\times$ } } & +0,30{\scriptsize \% } & 3,00{\scriptsize {\tiny $\times$ } } & 2,72{\scriptsize {\tiny $\times$ } } & +0,65{\scriptsize \% }& 1,28{\scriptsize {\tiny $\times$ } } & 0,31{\scriptsize {\tiny $\times$ } }\\
    \textit{svhn} &
    98.21{\scriptsize \% }& {\scriptsize ResNet-101} & +0,07{\scriptsize \% }& 0,70{\scriptsize {\tiny $\times$ } } & 1,10{\scriptsize {\tiny $\times$ } } & +0,06{\scriptsize \% }& 1,80{\scriptsize {\tiny $\times$ } } & 0,90{\scriptsize {\tiny $\times$ } } & +0,00{\scriptsize \% } & 1,00{\scriptsize {\tiny $\times$ }} & 1,00{\scriptsize {\tiny $\times$ }} & +0,02{\scriptsize \% }& 0,63{\scriptsize {\tiny $\times$ }} & 0,30{\scriptsize {\tiny $\times$ } }\\
    \textit{stl} &
    72.72{\scriptsize \% }& {\scriptsize GoogLeNet} & +4,02{\scriptsize \% }& 0,40{\scriptsize {\tiny $\times$ } } & 7,80{\scriptsize {\tiny $\times$ } } & +0,17{\scriptsize \% }& 3,15{\scriptsize {\tiny $\times$ } } & 3,00{\scriptsize {\tiny $\times$ } } & +0,02{\scriptsize \% } & 7,44{\scriptsize {\tiny $\times$ } } & 5,60{\scriptsize {\tiny $\times$ } } & +0,16{\scriptsize \% }& 0,71{\scriptsize {\tiny $\times$ } } & 0,94{\scriptsize {\tiny $\times$ } }\\
    \textit{mnist} &
    99.41{\scriptsize \% }& {\scriptsize MobileNet-V2} & +0,00{\scriptsize \% } & 1,00{\scriptsize {\tiny $\times$ } } & 1,00{\scriptsize {\tiny $\times$ } } &  +0,00{\scriptsize \% } & 1,00{\scriptsize {\tiny $\times$ } } &  1,00{\scriptsize {\tiny $\times$ } } & +0,00{\scriptsize \% } & 1,00{\scriptsize {\tiny $\times$ }} & 1,00{\scriptsize {\tiny $\times$ }} & +0,00{\scriptsize \% } & 1,00{\scriptsize {\tiny $\times$ } } &  1,00{\scriptsize {\tiny $\times$ } }\\
    \textit{gtsrb} & 
    96.56{\scriptsize \% }& {\scriptsize ResNet-101} & +0,41{\scriptsize \% }& 0,87{\scriptsize {\tiny $\times$ } } & 5,46{\scriptsize {\tiny $\times$ } } & +0,08{\scriptsize \% }& 3,15{\scriptsize {\tiny $\times$ } } & 2,98{\scriptsize {\tiny $\times$ } } & +0,02{\scriptsize \% } & 1,54{\scriptsize {\tiny $\times$ }} & 5,96{\scriptsize {\tiny $\times$ }} & +0,01{\scriptsize \% }& 1,03{\scriptsize {\tiny $\times$ } } & 0,90{\scriptsize {\tiny $\times$ } }\\
    \textit{food} &
    69.54{\scriptsize \% }& {\scriptsize ResNeXt-29} & +4,30{\scriptsize \% }& 0,34{\scriptsize {\tiny $\times$ } } & 3,40{\scriptsize {\tiny $\times$ } } & +0,19{\scriptsize \% }& 2,11{\scriptsize {\tiny $\times$ } } & 0,60{\scriptsize {\tiny $\times$ } } & +0,22{\scriptsize \% } & 2,73{\scriptsize {\tiny $\times$ }} & 1,39{\scriptsize {\tiny $\times$ } } & +0,30{\scriptsize \% }& 1,46{\scriptsize {\tiny $\times$ } } & 0,28{\scriptsize {\tiny $\times$ } }\\
    \textit{flowers} &
    59.19{\scriptsize \% }& {\scriptsize VGG-11} & +7,30{\scriptsize \% }& 0,10{\scriptsize {\tiny $\times$ } } & 3,75{\scriptsize {\tiny $\times$ } } & +0,00{\scriptsize \% } & 1,00{\scriptsize {\tiny $\times$ } } &  1,00{\scriptsize {\tiny $\times$ } } & +0,00{\scriptsize \% } & 1,00{\scriptsize {\tiny $\times$ }} & 1,00{\scriptsize {\tiny $\times$ }} & +2,10{\scriptsize \% }& 0,30{\scriptsize {\tiny $\times$ }} & 0,13{\scriptsize {\tiny $\times$ } }\\
    \textit{fashion} &
    95.39 {\scriptsize \% }& {\scriptsize PreActResNet-101} & +0,35{\scriptsize \% }& 0,31{\scriptsize {\tiny $\times$ } } & 2,48{\scriptsize {\tiny $\times$ } } & +0,07{\scriptsize \% }& 1,48{\scriptsize {\tiny $\times$ } } & 0,97{\scriptsize {\tiny $\times$ } } & +0,09{\scriptsize \% } & 4,44{\scriptsize {\tiny $\times$ } } & 1,19{\scriptsize {\tiny $\times$ } } & +0,04{\scriptsize \% }& 0,64{\scriptsize {\tiny $\times$ } } & 0,22{\scriptsize {\tiny $\times$ } }\\
    \textit{caltech} &
    52.74{\scriptsize \% }& {\scriptsize DenseNet-161} & +5,85{\scriptsize \% }& 0,59{\scriptsize {\tiny $\times$ } } & 1,65{\scriptsize {\tiny $\times$ } } & +0,19{\scriptsize \% }& 4,07{\scriptsize {\tiny $\times$ } } & 0,89{\scriptsize {\tiny $\times$ } } & +0,16{\scriptsize \% } & 8,12{\scriptsize {\tiny $\times$ }} & 0,96{\scriptsize {\tiny $\times$ }} & +0,08{\scriptsize \% }& 2,46{\scriptsize {\tiny $\times$ } } & 0,14{\scriptsize {\tiny $\times$ } }\\
    \midrule
    \textbf{Average} & & & +2,55{\scriptsize \% }& 0,53{\scriptsize {\tiny $\times$ } } & 3,27{\scriptsize {\tiny $\times$ } } & +0,10{\scriptsize \%}& 2,38{\scriptsize {\tiny $\times$ } } & 1,42{\scriptsize {\tiny $\times$ } } & +0,09{\scriptsize \%} & 3,66{\scriptsize {\tiny $\times$ }} & 2,33{\scriptsize {\tiny $\times$}} & +0,34{\scriptsize \% }& 1,12{\scriptsize {\tiny $\times$ } } & 0,45{\scriptsize {\tiny $\times$ }}\\
    \bottomrule
    \end{tabular}
\end{table*}


\begin{figure}
\centering
\includegraphics[width=8.5cm]{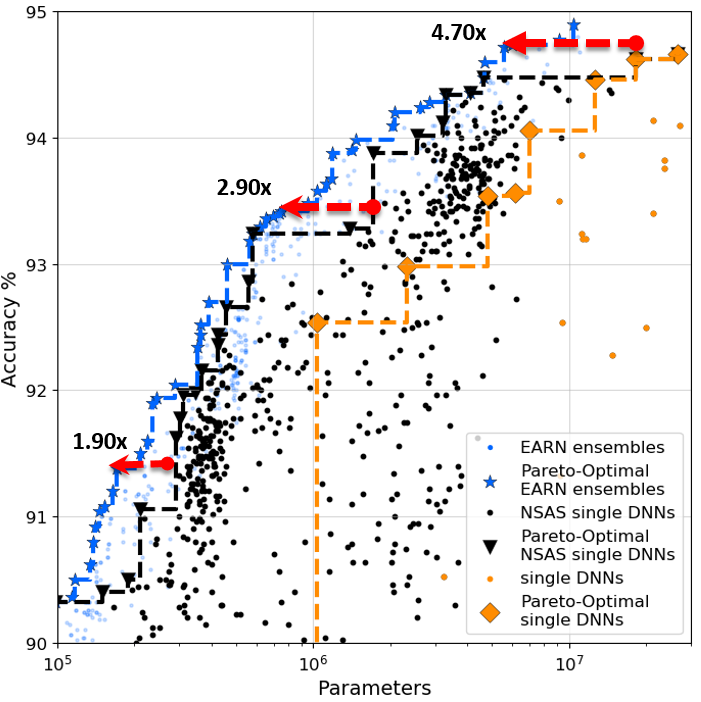}
\caption{EARN improving NSAS auto-generated DNNs.}
\label{fig:earn(nsas)}
\end{figure}

\textbf{Comparing EARN with state-of-the-art.}

We compare EARN to a Neural Architecture Search (NAS) approach \cite{scheidegger2019constrained} capable of delivering small, low-power, and high-performing neural network models targeting IoT devices. We refer to this approach as NSAS. 

We start by running NSAS to generate and train over 3k new DNN models on CIFAR-10 from a baseline of 32 diverse DNNs. The list of 32 DNNs, is described in the Experimental Section of this work. Afterward, we use EARN on the 3K NSAS models to generate even smaller and more accurate combined solutions. We execute EARN for 500 iterations.

We show the result of the experiment in Figure \ref{fig:earn(nsas)}. The x-axis reveals the size of the models, while the y-axis displays their test accuracy on CIFAR-10. In orange, the 32 DNNs baseline, in black the 3K DNNs from NSAS, and in blue, the EARN ensembles resulting from combining NSAS DNNs. Only top solutions with accuracy higher than 90\% on CIFAR-10 are displayed. As we can see in the plot, EARN improves NSAS solutions in many areas of the solution space, generating solutions with up to 4.70x fewer parameters with the same or higher accuracy. In addition, EARN populates the space with many more efficient solutions where NSAS fails to produce, for instance, around x=$10^6$. This evaluation illustrates how EARN can be used as a complement to NAS methods for model optimization problems by following the same methodology we apply in this experiment.

\subsection{Experimental Setup}
\label{sec:experimentalsetup}
The analyses we perform in this paper consider the following image classification data sets: cifar10, cifar100, svhn, stl10, mnist, fashion-mnist, gtsrb, food101, flowers102, and caltech256. We divide the test sets in half with an equal representation of samples per class, to create validation sets.


We consider 32 state-of-the-art DNN models: DenseNet with 121, 161, 169 and 201 depth; VGG with 11, 13, 16 and 19 layers of depth; PreActResNet with 18, 24, 50, 101 and 152 layers; ResNet with 18, 34, 50, 101 and 152 layers; ResNeXt29-2x64d, ResNeXt29-4x64d and ResNeXt29-32x4d; The MobileNet-V1 and MobileNet-V2; GoogLeNet. DPN with 26 and 29 depth; LeNet; PNASNet types A and B and finally the SENet-18.

This paper considers two different hardware platforms: A Nvidia Volta V-100 GPU, and a cluster composed of 48-cores nodes. Each node consists of 2x Intel Xeon Platinum 8160 24-cores processors.

An open-source\footnote{Omitted for blind review} Python (3.6) framework has been specifically developed for the experimental part of this paper. The framework makes it easy to build, and evaluate bagging/boosting/chain ensembles of ML classifiers. Evaluating an ensemble requires obtaining its classification accuracy, the total number of parameters, and its expected inference time.  

Since this paper performs a very large experimental campaign, we pre-compute the inference time of each DNN per each one of the 10 data test sets on the GPU device considering 128 sample batches. We also pre-compute the individual DNN predictions. Then, the expected inference time of an ensemble is estimated in the following way: Since we know the inference time of all the classifiers in the ensemble on the Volta GPU with batches of 128, the estimated ensemble inference time is the sum of the inference times of the each of the classifiers participating in the inference. The classification accuracy is also estimated from the individual predictions of the models forming the ensemble. This approach allows for a fast evaluation rate, and it was only used with the GPU platform. The CPU results correspond to real ensemble evaluations on PyTorch 1.2.0.

\section{Related Work}

Most of the early works on automatic ensemble generation combine all trained base learners to construct the ensemble. For instance, a previous approach~\cite{sylvester2005evolutionary} proposes a single-objective evolutionary algorithm called EVEN. It modifies a set of weights to maximize the weighted prediction of all trained learners on the validation set. With EVEN, the performance of the ensemble is larger than combining the learners with a voting protocol and also larger than the best learner in the ensemble.

Selecting a subset of the initial pool of trained learners not only reduces the storage requirements and prediction response, but it could lead to a better generalization performance and similar predicting results \cite{zhou2002ensembling}. This technique is often referred to in the literature as \textit{ensemble pruning}. Most contemporary work on ensemble generation adopts pruning for the generation of ensembles according to two objectives: \textit{prediction performance} and \textit{diversity} \cite{1716772,4041439,bian2019does,zaidi2020neural,chandra2004divace,nan2012diversity}. Encouraging diversity in the ensemble can be seen as a regularization method that makes it possible to achieve better generalization. When the members of the ensemble are accurate and diverse, the ensemble prediction performance is enhanced. 

In addition, \textit{prediction performance} and \textit{ensemble size} are two other objectives that are jointly optimized for the generation of efficient ensembles\cite{qian2015pareto,zhang2010creating,chen2009efficient}. Previous work~\cite{qian2015pareto} introduces PEP, a mutation-only evolutionary algorithm guided by a bi-objective fitting function and combined with a local search operator. PEP outperforms in terms of test error similar ensemble pruning methods with smaller-sized homogeneous ensembles. 
Another approach~\cite{wang2018quit} developed QWYC, a greedy algorithm that defines an execution order for homogeneous learners, and sets threshold values for early-exit conditions in binary classification tasks, thus accelerating the inference with minimal prediction degradation. In their work, they too consider homogeneous ML models, and they minimize classification error and the activated ensembles during inference.

In this work, we introduce EARN, an automatic method that generates optimal ensembles for CPU and GPU devices according to three objectives: classification accuracy, latency time, and memory footprint. EARN combines the members in the ensemble with multiple ensemble strategies we combine mutation with crossover operations to accelerate the generation. To the best of our knowledge, this is the first work that automatically generates optimal ensembles by combining DL models with multiple ensemble strategies.

\section{Conclusions}

We use ensemble methods as a technique for generating more efficient DL solutions. We propose to combine multiple ensemble strategies to increase prediction performance and reduce computational costs. Since an exhaustive search for the optimal ensemble configuration is unfeasible we introduce EARN, an evolutionary algorithm that optimizes the ensembles according to 3 objectives with very few evaluations.

We analyze the ensembles generated by EARN on 10 image datasets, demonstrating great improvements in classification accuracy, prediction response time, and size. EARN generates ensembles with +7.30\% accuracy increase, $4.70\times$ speedup on a GPU device, or $7.70\times$ size reduction.

Our approach is combined with a recently proposed Neural Architecture Search (NAS) method for constructing efficient predictors. We show how EARN can be applied to the models generated by the NAS approach to produce even smaller and more accurate models.

\bibliographystyle{ijcai/named}
\bibliography{bibliography}

\begin{thebibliography}{}

\bibitem[\protect\citeauthoryear{Arjun and Xin}{2004}]{chandra2004divace}
Chandra Arjun and Yao Xin.
\newblock Divace: Diverse and accurate ensemble learning algorithm.
\newblock In {\em International Conference on Intelligent Data Engineering and
  Automated Learning}, pages 619--625. Springer, 2004.

\bibitem[\protect\citeauthoryear{Breiman}{1996}]{breiman1996bagging}
L.~Breiman.
\newblock Bagging predictors.
\newblock {\em Machine learning}, 24(2):123--140, 1996.

\bibitem[\protect\citeauthoryear{Breinan}{2001}]{breiman2001random}
L.~Breinan.
\newblock Random forests.
\newblock {\em Machine learning}, 45(1):5--32, 2001.

\bibitem[\protect\citeauthoryear{Chao \bgroup \em et al.\egroup
  }{2015a}]{qian2015pareto}
Qian Chao, Yu~Yang, and Zhou Zhi-Hua.
\newblock Pareto ensemble pruning.
\newblock In {\em AAAI}, pages 2935--2941, 2015.

\bibitem[\protect\citeauthoryear{Chao \bgroup \em et al.\egroup
  }{2015b}]{qian2015subset}
Qian Chao, Yu~Yang, and Zhou Zhi-Hua.
\newblock Subset selection by pareto optimization.
\newblock In {\em Advances in Neural Information Processing Systems}, pages
  1774--1782, 2015.

\bibitem[\protect\citeauthoryear{Eckart \bgroup \em et al.\egroup
  }{2001}]{zitzler2001spea2}
Zitzler Eckart, Laumanns Marco, and Thiele Lothar.
\newblock Spea2: Improving the strength pareto evolutionary algorithm.
\newblock {\em TIK-report}, 103, 2001.

\bibitem[\protect\citeauthoryear{Evgenii and Ilia}{2015}]{sopov2015self}
Sopov Evgenii and Ivanov Ilia.
\newblock Self-configuring ensemble of neural network classifiers for emotion
  recognition in the intelligent human-machine interaction.
\newblock In {\em 2015 IEEE Symposium Series on Computational Intelligence},
  pages 1808--1815. IEEE, 2015.

\bibitem[\protect\citeauthoryear{Freund}{2001}]{freund2001adaptive}
Y.~Freund.
\newblock An adaptive version of the boost by majority algorithm.
\newblock {\em Machine learning}, 43(3):293--318, 2001.

\bibitem[\protect\citeauthoryear{Graning \bgroup \em et al.\egroup
  }{2006}]{1716772}
L.~Graning, J.~Yaochu, and B.~Sendhoff.
\newblock Generalization improvement in multi-objective learning.
\newblock In {\em The 2006 IEEE International Joint Conference on Neural
  Network Proceedings}, pages 4839--4846, 2006.

\bibitem[\protect\citeauthoryear{Hansen and Salamon}{1990}]{58871}
L.~K. Hansen and P.~Salamon.
\newblock Neural network ensembles.
\newblock {\em IEEE Transactions on Pattern Analysis and Machine Intelligence},
  12(10):993--1001, 1990.

\bibitem[\protect\citeauthoryear{Hiroshi}{2019}]{inoue2019adaptive}
Inoue Hiroshi.
\newblock Adaptive ensemble prediction for deep neural networks based on
  confidence level.
\newblock In {\em The 22nd International Conference on Artificial Intelligence
  and Statistics}, pages 1284--1293. PMLR, 2019.

\bibitem[\protect\citeauthoryear{Huaxiang and Jing}{2010}]{zhang2010creating}
Zhang Huaxiang and Lu~Jing.
\newblock Creating ensembles of classifiers via fuzzy clustering and
  deflection.
\newblock {\em Fuzzy sets and Systems}, 161(13):1790--1802, 2010.

\bibitem[\protect\citeauthoryear{Jared and
  Nitesh}{2005}]{sylvester2005evolutionary}
Sylvester Jared and Chawla Nitesh.
\newblock Evolutionary ensembles: Combining learning agents using genetic
  algorithms.
\newblock In {\em AAAI Workshop on Multiagent Learning}, pages 46--51, 2005.

\bibitem[\protect\citeauthoryear{Kalyanmoy \bgroup \em et al.\egroup
  }{2002}]{deb2002fast}
Deb Kalyanmoy, Pratap Amrit, Agarwal Sameer, and Meyarivan Tamt.
\newblock A fast and elitist multiobjective genetic algorithm: Nsga-ii.
\newblock {\em IEEE transactions on evolutionary computation}, 6(2):182--197,
  2002.

\bibitem[\protect\citeauthoryear{Koza and Koza}{1992}]{koza1992genetic}
John~R Koza and John~R Koza.
\newblock {\em Genetic programming: on the programming of computers by means of
  natural selection}, volume~1.
\newblock MIT press, 1992.

\bibitem[\protect\citeauthoryear{Li \bgroup \em et al.\egroup
  }{2012}]{nan2012diversity}
Nan Li, Yu~Yang, and Zhou Zhihua.
\newblock Diversity regularized ensemble pruning.
\newblock pages 330--345, Berlin, Heidelberg, 2012. Springer Berlin Heidelberg.

\bibitem[\protect\citeauthoryear{Nojima and Ishibuchi}{2006}]{4041439}
Y.~Nojima and H.~Ishibuchi.
\newblock Designing fuzzy ensemble classifiers by evolutionary multiobjective
  optimization with an entropy-based diversity criterion.
\newblock In {\em 2006 Sixth International Conference on Hybrid Intelligent
  Systems}, pages 59--59, 2006.

\bibitem[\protect\citeauthoryear{Scheidegger \bgroup \em et al.\egroup
  }{2019}]{scheidegger2019constrained}
Florian Scheidegger, Luca Benini, Costas Bekas, and Cristiano Malossi.
\newblock Constrained deep neural network architecture search for iot devices
  accounting for hardware calibration.
\newblock In {\em Advances in Neural Information Processing Systems}, pages
  6056--6066, 2019.

\bibitem[\protect\citeauthoryear{Serena \bgroup \em et al.\egroup
  }{2018}]{wang2018quit}
Wang Serena, Gupta Maya, and You Seungil.
\newblock Quit when you can: Efficient evaluation of ensembles with ordering
  optimization.
\newblock {\em arXiv preprint arXiv:1806.11202}, 2018.

\bibitem[\protect\citeauthoryear{Sheheryar \bgroup \em et al.\egroup
  }{2020}]{zaidi2020neural}
Zaidi Sheheryar, Zela Arber, Elsken Thomas, Holmes Chris, Hutter Frank, and Teh
  Yee.
\newblock Neural ensemble search for performant and calibrated predictions.
\newblock {\em arXiv preprint arXiv\:2006.08573}, 2020.

\bibitem[\protect\citeauthoryear{Trevor \bgroup \em et al.\egroup
  }{2009}]{hastie2009multi}
Hastie Trevor, Rosset Saharon, Zhu Ji, and Zou Hui.
\newblock Multi-class adaboost.
\newblock {\em Statistics and its Interface}, 2(3):349--360, 2009.

\bibitem[\protect\citeauthoryear{Viola and Jones}{2004}]{viola2004robust}
Paul Viola and Michael Jones.
\newblock Robust real-time face detection.
\newblock {\em International journal of computer vision}, 57(2):137--154, 2004.

\bibitem[\protect\citeauthoryear{Yan \bgroup \em et al.\egroup
  }{2009}]{chen2009efficient}
Chen Yan, Oliver~Dean S, and Zhang Dongxiao.
\newblock Efficient ensemble-based closed-loop production optimization.
\newblock {\em SPE Journal}, 14(04):634--645, 2009.

\bibitem[\protect\citeauthoryear{Yi \bgroup \em et al.\egroup
  }{2006}]{zhang2006ensemble}
Zhang Yi, Burer Samuel, and Street~W Nick.
\newblock Ensemble pruning via semi-definite programming.
\newblock {\em Journal of machine learning research}, 7(Jul):1315--1338, 2006.

\bibitem[\protect\citeauthoryear{Yijun and Huanhuan}{2019}]{bian2019does}
Bian Yijun and Chen Huanhuan.
\newblock When does diversity help generalization in classification ensembles?
\newblock {\em arXiv preprint arXiv:1910.13631}, 2019.

\bibitem[\protect\citeauthoryear{Yoav and Robert}{1996}]{freund1996experiments}
Freund Yoav and Schapire Robert.
\newblock Experiments with a new boosting algorithm.
\newblock In {\em icml}, volume~96, pages 148--156. Citeseer, 1996.

\bibitem[\protect\citeauthoryear{Zhihua \bgroup \em et al.\egroup
  }{2002}]{zhou2002ensembling}
Zhou Zhihua, Wu~Jianxin, and Tang Wei.
\newblock Ensembling neural networks: many could be better than all.
\newblock {\em Artificial intelligence}, 137(1-2):239--263, 2002.

\bibitem[\protect\citeauthoryear{Zitzler and
  Thiele}{1999}]{zitzler1999multiobjective}
Eckart Zitzler and Lothar Thiele.
\newblock Multiobjective evolutionary algorithms: a comparative case study and
  the strength pareto approach.
\newblock {\em IEEE transactions on Evolutionary Computation}, 3(4):257--271,
  1999.

\end{thebibliography}

\end{document}